\pdfoutput=1

\documentclass[11pt]{article}

\usepackage[final]{acl}
\usepackage{balance}
\usepackage{url}
\usepackage{times}
\usepackage{latexsym}
\usepackage{xspace}
\usepackage{booktabs}
\usepackage{colortbl}
\usepackage[shortlabels]{enumitem}

\usepackage[T1]{fontenc}

\usepackage[utf8]{inputenc}

\usepackage{microtype}

\usepackage{graphicx} 
\newcommand{\bname}{\ensuremath{\mathsf{TimelineQA}}\xspace}
\newcommand{\llog}{{lifelog}}
\newcommand{\llogs}{{lifelogs}}

\definecolor{mygreen}{RGB}{34, 139, 34}
\definecolor{myred}{RGB}{209, 0, 86}
\definecolor{myblue}{RGB}{110, 245, 227}

\definecolor{nav}{RGB}{0,0,128}

\title{TimelineQA: A Benchmark for Question Answering over Timelines}

\author{Wang-Chiew Tan, Jane Dwivedi-Yu, Yuliang Li, \\
{\bf Lambert Mathias, Marzieh Saeidi\textsuperscript{*}, Jing Nathan Yan\textsuperscript{+}, and
Alon Y. Halevy} \\
Meta~~~~Cornell University$^+$\\
  \texttt{\{wangchiew,janeyu,yuliangli,lambert,ayh\}@meta.com}\\ \texttt{marzieh.saeidi@googlemail.com}\textsuperscript{*} ~~\texttt{jy858@cornell.edu}\textsuperscript{+}}

\begin{document}

\maketitle

\begin{abstract}
Lifelogs are descriptions of experiences that a person had during their life. Lifelogs are created by fusing data from the multitude of digital services, such as online photos, maps,  shopping and content streaming services. Question answering over lifelogs can offer personal assistants a critical resource when they try to provide advice in context.  However, obtaining answers to questions over lifelogs is beyond the current state of the art of question answering techniques for a variety of reasons, the most pronounced of which is that lifelogs combine free text with some degree of structure  such as temporal and geographical information.  

We create and publicly release \bname\footnote{Code and data available at 
\url{https://github.com/facebookresearch/TimelineQA}},
a benchmark for accelerating progress  on querying lifelogs. \bname generates lifelogs of imaginary people. The episodes in the lifelog range from major life episodes such as high school graduation to those that occur on a daily basis such as going for a run. We describe a set of experiments on \bname\ with several state-of-the-art QA models. Our experiments reveal that for atomic queries, an extractive QA system significantly out-performs a state-of-the-art retrieval-augmented QA system. For multi-hop queries involving aggregates, we show that the best result is obtained with a state-of-the-art table QA technique, assuming the ground truth set of episodes for deriving the answer is available.

\end{abstract}

\section{Introduction}

The promise of augmented reality (AR) glasses has renewed interest in building personal assistants that are capable of being with us at all times of the day. In order for such assistants to be useful, they need to have detailed knowledge about the user, including their past experiences, preferences, habits and goals in the spirit of systems such as Memex~\cite{DBLP:journals/theatlantic/Bush45} and MyLifeBits~\cite{DBLP:journals/cacm/GemmellBL06}.  A lot of that knowledge already is implicitly present in the digital data that people generate by interacting with a myriad of online services such as photos, maps, health apps, shopping and content streaming.  A lifelog is a private and secure database that contains a set of episodes from the user's past that are gleaned from these data sources and in the future from smart glasses. The lifelog is completely under the control of the user, and only they can decide if and when to share fragments of it as they see beneficial. For example, they may share past dining experiences with an assistant when trying to choose an item from a menu, or past movie preferences with a friend when trying to decide which movie to watch together.   

In addition to issues relating to privacy, lifelogs raise two main classes of challenges. The first is to infer meaningful episodes from the raw data. For example, such an inference module would take as input a set of photos and output an episode such as {\em visited Venice for 7 days}, or {\em celebrated a birthday party with friends}. The second challenge, which is the subject of this paper is to answer questions over the lifelog, such as {\em when did I go to Tokyo}, {\em what did I eat on my second night in Paris}, or {\em how many times did I go to the dentist last year}.

Question answering is challenging because the data contains a combination of text and structure. The episodes themselves are described as text (and may also contain images and video), but each episode is associated with a time and location. For example, in order to answer a query such as {\em where did I take my mom when she visited Seattle}, the system first needs to figure out when mom visited Seattle and then look for episodes within that time interval. Other questions may require counting or reasoning over sets of episodes, similar to challenges raised in~\cite{DBLP:conf/acl/ThorneYSS0H20}. 

This paper describes \bname, a benchmark for querying lifelogs. The benchmark includes a generator that produces lifelogs for imaginary people with different personas (e.g., age, gender, education and family status). Each lifelog includes episodes drawn from a variety of activities, ranging from significant activities (e.g.,  going on a trip or getting married) to more daily activities (e.g., cooking dinner or going to the doctor). For each lifelog, the benchmark creates a set of question/answer pairs, specified in English. 

Naturally, real lifelogs are complex and extremely diverse and are challenging to generate synthetically. 
Our main contribution is a benchmark for QA systems over lifelog data of different sizes. The goal of the benchmark is not to represent people's lives in their full complexity or diversity, but to offer a sufficiently rich set of lifelogs that already exposes the challenges involved in question answering (QA). We show some snippets of our generated lifelogs Section~\ref{sec:lifelogexample}. As our QA techniques improve, the benchmark will be enriched to include more real life complexities. 

We describe a set of experiments demonstrating that current SOTA QA techniques fall short of adequate performance on lifelogs. We experimented with extractive \cite{karpukhin-etal-2020-dense} and RAG \cite{lewis2020rag} QA systems on atomic queries. Somewhat surprisingly, even after fine-tuning, the generative RAG QA system still lags behind the extractive system for question-answering.
In addition, we ran a Tapex~\cite{DBLP:conf/iclr/LiuCGZLCL22}, a table QA model and BART~\cite{DBLP:conf/acl/LewisLGGMLSZ20} for complex queries over \bname. Our experiments reveal that the best performing system, Tapex, only scores 59.0\%, assuming that the subset of episodes needed to compute the answer is known. 

\section{Related work}

The idea of creating a repository that captures all the knowledge about a person's life dates back to Vannevar Bush's vision of the Memex System~\cite{DBLP:journals/theatlantic/Bush45}. ~\citet{DBLP:journals/cacm/GemmellBL06} describes the MyLifeBits System that implemented the vision with the technology available in the late 1990's, and they used simple keyword search with the help of an SQL database to query its contents. \citet{DBLP:conf/mir/AlamGG22} describes a more recent project on creating lifelogs, and the  Solid Project (\citet{DBLP:conf/www/MansourSHZCGAB16}) takes an even more radical approach, suggesting that all of the user's data be stored in a {\em data pod} and that applications be redesigned to access it from the pod. Since the early years, the promise of personal agents has increased since data storage has become cheaper and ubiquitous, we anyway generate many more digital breadcrumbs with services we use on a daily basis, and AI techniques have become much better at analyzing text and image content.

The design of our benchmark was inspired by the Clevr benchmark for evaluating visual query answering systems~\cite{DBLP:conf/cvpr/JohnsonHMFZG17}. Like Clevr, we design a space of possible questions that can be asked and then generate synthetic datasets where we know the answer to each questions posed. 

There is a rich body of work on query answering. The ones closest to our work are on multi-hop queries~\cite{multihopsurvey2022} and neural databases~\cite{DBLP:conf/acl/ThorneYSS0H20}. In addition to queries that can be answered from a single episode in a lifelog, \bname includes  more complex queries that require combining information from multiple episodes in a lifelog. This is similar to work on QA over long documents~\cite{khashabi-etal-2018-looking}. However, the length of a lifelog is typically much greater than any existing benchmark or experimental dataset to the best of our knowledge. A typical lifelog can contain between 15M to 78M entries on average, where each entry contains about 8--9 tokens on average. 
Furthermore, \bname\ queries can also contain aggregates (e.g., max, sum, average). Neural databases considers the problem of answering aggregate queries over text data of arbitrary size, but it does not address the temporal aspects that are critical to queries over lifelogs. 


\section{Lifelogs}
\begin{figure*}[htb]

  \begin{center}
    \includegraphics[width=\linewidth]{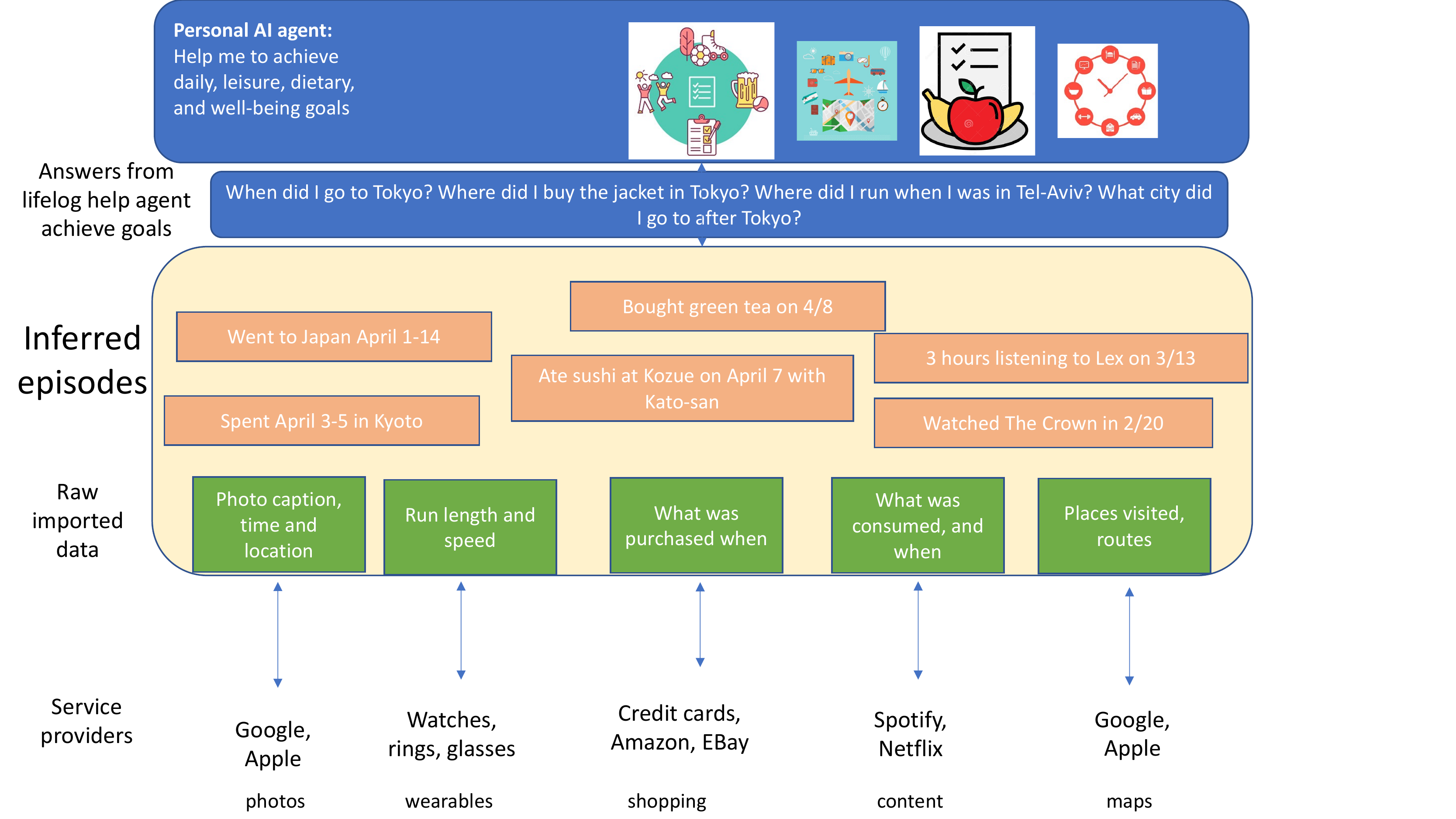}
  \end{center}
  \vspace{-0.3cm}
  \caption{Lifelogs import meta-data from a set of external services. A set of inference models deduces higher level episodes from the raw imported data. Episodes have a start and end time, and often a location. Question answering uses the raw and the deduced data. }
  \label{fig:lifelog}
\end{figure*}

A \llog\ includes any kind of experience that a user recorded digitally (see Figure~\ref{fig:lifelog}). We model experiences as {\em episodes} in the lifelog, and every episode is associated with a start/end time and start/end location, if those are known. Episodes are captured  via photos or videos, smart watches (e.g., exercise and sleep tracking),  mapping services (e.g., routes and visits), documents that have been explicitly stored (e.g., passport), or notes that the user takes describing their subjective experiences. 
A \llog\ is completely private and accessible only to the user. She can share slices of her \llog\ if and when there's value in doing so (e.g., getting better service from a sales person). 

Episodes are typically activities that the user was involved in, such as celebrating a holiday, going on a trip, going for a run or a bike ride, physical therapy, seeing fireworks or watching a movie. 
Episodes in the \llog\ can either be done by the owner of the \llog\ or by someone in their family or circle of acquaintances, e.g., mom moving to Seattle, sister getting married, having one's air-conditioning fixed, or being told something by a friend.
In addition to time and location, episodes may have attributes, such as  who was involved, the distance and speed of a run, or the name of a product that was purchased.  Some of these attributes may be modeled explicitly in the \llog\ if they're easy to extract,  and others may remain in the raw text or image and found at query time. 

Lifelogs are meant to be built with as little friction as possible from users. Hence, as shown in Figure~\ref{fig:lifelog}, the data is imported from the external services into the lifelog as raw data. Some raw data already describes episodes (e.g., purchase or content consumption episodes). Other episodes are then inferred by analyzing and fusing multiple pieces of raw data (e.g., a trip, or a meal with friends).  Of course, the inference step is a best-effort one, which means that some questions may still be impossible to answer and in some cases the QA system will point the user back to data that contains the answer (e.g., what did we eat on my daughter's birthday). Questions are answered based on the text and structured data describing all the episodes in the lifelog.   

Our work concerns question answering after the inference of episodes has been done.  Hence, formally a \llog\ is a collection of episodes, each one associated with their start/end time and location:  time-and-space boxed episodes.  Each episode contains some text and possibly pointers to external raw data. 
Note that episodes can be nested within other episodes.

\subsection{A classification of questions}
To understand the breadth and types of questions users may want to ask of lifelogs,  we crowdsourced the task of writing down questions over their potential lifelogs to 7 people. We also asked for the categories of their questions. We obtained a total of about 600 questions. We analyzed the categories and organized them into 13 topics (e.g., life milestones, travel, daily activities) as described in Table~\ref{tbl:question_categories} in the appendix. After this, we asked (again) each contributor to write a few questions they would ask on each of the 13 topics. 

Based on a qualitative analysis of these questions, we observe that the queries can be classfied as follows. We use the term query and question interchangeably. 

\medskip
\noindent
{\bf Atomic questions:}
An atomic query, which is the most common type, asks for some attribute of an episode. Examples include: 
\vspace{-2mm}
\begin{itemize}
    \item When did my mom have a knee operation?
    \vspace{-2mm}
    \item What's the name of the company that repaired my A/C?
    \vspace{-2mm}
    \item What's the name of my daughter's first-grade teacher?
    \vspace{-2mm}
\end{itemize}

An atomic query is one that can be answered by a {\em single} episode. 
The answer to an atomic query can either be directly explicit in the text of the episode (e.g., when), or requires inference from the text (e.g., who fixed the A/C). 
For example, if an episode describes ``{\em 08/01/2022: John was here. He fixed the AC this morning.}'', then the respective answers to the questions are ``08/01/2022'' and ``John''. In principle, an answer may also be a link to a photo that may contain the information asked by the user, though \bname is currently limited to questions that can be answered after the inference of episodes is done.  
Finally, some answers may require a bit of derivation. For example, when is my sister's 40th birthday could be derived from the episode describing her birth.

\medskip
\noindent
{\bf Complex queries -- multi-hop:} 
The answer to a multi-hop query is
formed by combining data from multiple episodes. Hence, oftentimes, multi-hop queries require identifying a set of episodes in the timeline. For example, \textit{Where did we eat great Indian food on our way to Vancouver?} would require identifying episodes involving the trek to Vancouver and eating Indian food. Other examples of multi-hope queries are:
\vspace{-2mm}
\begin{itemize}
\item What places did I visit when my mom came to visit Seattle?
\vspace{-2mm}
\item  Show me photos of the car damage I had after the accident
\end{itemize}

\smallskip
\noindent
{\bf Complex queries -- aggregates:}  
These questions (known as {\em aggregation queries} in SQL) consider a set of episodes and compute an aggregate on them. For example:
\vspace{-2mm}
\begin{itemize}  
\item How many times did I visit the dentist this year?
\vspace{-2mm}
\item How many miles did I bike this year?
\end{itemize}
In some cases, the aggregation may be combined with another condition, such as {\em How many calories did I burn on my last two rides?} or {\em When did I last ride 40 miles or more in a day?} 

\medskip
\noindent
{\bf Temporal queries:}
Because of the nature of personal timelines, many of the questions that arise are temporal ones. In addition to atomic and complex queries, we identified temporal queries that may be atomic or complex. Examples of atomic queries that are temporal are those whose answer is the time of an episode, such as ``{\em When did I pay my car insurance?}''
In general, temporal questions may require more sophisticated reasoning about time, such as finding the length of a life event or the time between episodes, e.g., {\em How long was my break between leaving my last job and starting my current job?} or reasoning about the sequence of occurrence ``{\em Did I go to Spain before Italy?}''. In our crowdsourced query collection, temporal queries were mostly atomic {\em when} queries or implicit subgoals of more complex queries, e.g., ``{\em when was the last time I visited the dentist?}''

\subsection{The goals of the benchmark}

The above classification of questions highlights some of the challenges that will arise in query answering over \llogs. The first challenge is typical for query answering---the disparity between the terms that are used in the query versus the language used in the \llog\ itself. For example, a user might ask when they had a drink with a particular friend, while the \llog\ may say that they went to a bar before dinner. In the \llog\ context the challenge can also require multi-modal reasoning, because the only item in the \llog\ might be a photo from a bar. As another example, users may refer to more aggregate terms than what's in the \llog. For example, the user may ask how much they spent on utilities last month, while the \llog\ has individual utility bills, but the system may not be aware of which bills are considered utility bills. We expect that query answering over \llogs\ will benefit from advances in the broader field of query answering and therefore this is not a focal point of our benchmark. 

The second set of challenges involves the interplay between the structure that the \llog\ supports and the linguistic reasoning. 
For example, the \llog\ may store the duration of every exercise you made, but answering the query on how long did you exercise every day for the past month is more challenging. 
Another complex example is in the context of multi-hop questions. If a user asks when was the first time she traveled to Tokyo, the system needs to find all instances of the user's travel to Tokyo and then return the first one.
Reasoning about such temporal relations is an area of weakness for QA algorithms today. This aspect of query answering is critical to \llogs\ and therefore we design our benchmark to evaluate these challenges.

Specifically, we would like our benchmark to push the limits on the interaction between structure and language in query answering. To that end, our benchmark is designed to be able to vary a few variables, including the complexity of the questions, the size and contents of the \llogs, and the types of data that are in the \llog, including the complexity of life episodes the user has, how verbose the user is  (i.e., do they log only their major experiences or also many minutiae episodes). 


\section{Creating \llogs\ in \bname}

Since we believe that \bname\ is the first in a series of \llog\ benchmarks, we explain here in some detail how it is built. 
A \llog\ is a set of episodes in the life of a person. Our goal is to create
\llogs\ that contain a good range of experiences 
that a
person may have in life and sufficient to begin benchmarking the performance of QA systems on \llogs. To collect a broad set of typical episodes, we started
with a detailed set of episode categories described in
Coelition\footnote{https://coelition.org/business/resources/visualising-life/},
a site that {\em provides technology and expert advice for data collected about
people on the Internet of Things}, and distilled them into the categories shown
in Table~\ref{table:categories}. The categories of episodes range from life
episodes (e.g., being born, going to college), episodes that happen a few times
a year (e.g., trips) to those that happen on a weekly or daily basis (e.g.,
meeting friends or cooking). 
The timescales and examples in Table~\ref{table:categories} coincided broadly with the types and categories of questions we obtained from our crowdsourced task. See Table~\ref{tbl:question_categories} in Appendix~\ref{sec:categories}.

\smallskip
\noindent
{\bf Creating a persona~}
The process of building a lifelog begins with creating a persona which includes the skeletal details of a person's life, including when and where they were born, their gender, their educational and professional history, their family members and some
of their preferences and hobbies. We first generate a birthdate, which must be between 18-75 years old at the time of generation. We randomly select a gender and a name from a dictionary of names. We then proceed to create their educational and professional history, family members, preferences and hobbies. These are generated via a model that
depends on several probability distributions of episodes. We note that while
the personas we create are quite varied, we do not claim that they represent a
diversity in any social sense. The diversity we do build in is limited: age,
gender, locations, professions. Clearly, in order to achieve robust query
answering on \llogs\ we need to consider many other kinds of diversity
(culture, non-typical episodes and scenarios), but we believe that the
benchmark as is already poses many important challenges.

\begin{table*}[tbh]

\centering
\small
\begin{tabular}{|c|c|}
\hline
{\bf Time scale} & {\bf Examples}  \\ \hline  
  Lifetime            &     birth, educational milestones, marriage, divorce, jobs \& relocation              \\ \hline
  Annual             &    travel, medical and dental checkups              \\ \hline
   Monthly            &   pet care (e.g., grooming)                 \\ 
   \hline
   Weekly            &   baking, cooking, dating, hobbies, buying groceries                \\ 
   \hline
   Daily            &   eating meals, talking with friends, exercising, consuming content (books, movies)                 \\ 
   \hline
\end{tabular}
\normalsize
\caption{\small Types of episodes in \bname. Episodes are divided into several time scales. The generator creates episodes in successive time scales, starting from lifetime events. }
\label{table:categories}
\vspace{-3mm}
\end{table*}

\smallskip
\noindent
{\bf Creating episodes~}
Once a persona is created, we begin creating episodes starting from the day the person was 18 years old to the present year.
We first create episodes in the \llog\ for life events, such as birth, educational phases, starting and ending jobs, marriage(s) and having
children. We then proceed to generate episodes at different levels of granularity based on the timescales (annual, monthly, weekly, daily) as shown in Table~\ref{table:categories}. For example, for annual episode types, we create annual health checkups episodes and yearly trips. For monthly episodes, we generate pet grooming episodes and some examples of weekly and daily episodes are baking/cooking, grocery shopping, catching up with friends or news. These episodes are generated as described in Table~\ref{table:categories}. These episodes are generated based on a predefined probability distribution which can be modified. 

Some of the episodes we create are {\em super episodes}, which involves
sub-episodes that depict events of finer granularity. 
For example, a multi-day travel or trip episode
will be
broken down to movements between different destinations, and the itinerary for
every single day and special episodes that happened in each day.  
The descriptions of episodes are generated by instantiating templates that 
we specify. Every episode is associated with a set of alternative templates and 
a template is randomly picked and instantiated for a given episode to be created.  
Since the templates are fixed, the descriptions generated may not offer the variety 
in descriptions we expect from a general population. 
We are in the process of incorporating the use of language models to generate episode descriptions as yet another alternative.
However, it is interesting to understand 
what limitations on QA such a benchmark already exposes 
with templated descriptions.

More variations in the episode activities can be added to the lifelog generator to more closely reflect the categories we find in the Coelition and also what we crowdsourced (Table~\ref{tbl:question_categories}). We leave this for future work.

\paragraph{Consistency through constraints:} To ensure
more consistency, we keep track of the attributes of every single day in one's
life. For example, the probability of certain episodes can change drastically
if a person is on a trip or in the process of getting married.
In \bname, constraints can be specified to prevent inconsistencies from occurring.
For example, since it is much less likely that one bakes or has an annual 
dental checkup while traveling, 
we can explicitly state that
these episodes should be mutually exclusive in \bname.
If an episode is to be created on a certain day, 
\bname\ checks that it is mutually exclusive to any existing episode applicable to that day
before creating the new episode.

\paragraph{Generating questions and answers:} 

Every lifelog,  $\cal D$, in \bname\ is associated with a set of
question/answer pairs $(Q,A)$, where $Q$ is a natural language question over
$\cal D$ and $A$ is the correct answer to it.  In order to ensure that we can
create a variety of questions that are meaningful on a particular \llog\ $\cal
D$ and that we know the correct answers to them, the process of creating begins
by creating  a logical representation of the episodes in the \llog\ and of the
questions and the answers, and then turning them into natural language. The
natural language of the questions and answers are created by instantiating a
few templates for every episode type. Because we use templates,  \bname\
clearly lacks the richness of linguistic variation, but as noted previously, dealing
with linguistic variation is not the focus of this benchmark.

We generate questions and answers for each lifelog in two steps: atomic
questions and complex questions. Since atomic questions are ones whose answer is
contained in a single episode in the lifelog, we can create them at the same time the episode is created.  
 For example, if the
episode is {\em I went to a Japanese restaurant with Sarah on October 7th and
ate sushi}, then we would generate questions of the form: {\em when did I have
Japanese food?}, {\em when did I meet Sarah?}, and {\em where did I eat on
October 7th?} For each single episode, we create {\em what}, {\em where}, {\em
when} and {\em who} questions as appropriate along with the corresponding answers.

Complex questions are ones that either rely on a set of facts in the lifelog,
such as, {\em how many times did I go to London?} and {\em where did I spend the
first night in Tokyo?} or require combining multiple facts as in multi-hop
questions such as, {\em which restaurants did I go to during my trip to New
York?} To create such question/answer pairs easily, we create a database of the
logical representation of all the episodes in the lifelog. We then consider a
set of query templates and check whether the template can be instantiated on
that database. Examples of templates we consider are:
\vspace{-2mm}
\begin{itemize}
    \item How many times did I X?
    \vspace{-2mm}
    \item When was the first/last time I X?
    \vspace{-2mm}
    \item Did I go to X before I went to Y?
    \vspace{-2mm}
    \item How many times did I do X when I was at Y?
\end{itemize}
Since we have all the episodes, we can compute the answers to these questions
correctly. 

\paragraph{Size and density:}
Lifelogs of different sizes can be created with \bname. The user specifies a year
and duration parameter, and this will determine the length of the lifelog to generate. 
 For example, if the year is 2023 and the duration is 5, then 5 years of episodes from 2018 to 2023, including lifetime episodes, will be created. Lifetime episodes such as birth and college education may occur outside those 5 years.

The user can also specify the density (sparse, medium, or dense) of episodes to generate in the lifelog.
The variations in density are used to mimic that different users log their
life events at different frequencies.
For example, if the generator is called with the ``sparse'' parameter, then 
the probabilities of 
generating daily/weekly/monthly episodes will be much lower than the case 
when the generator is called with the ``dense'' parameter. 

\subsection{Example lifelog}
\label{sec:lifelogexample}
An example snippet of our generated lifelog and sample question and answer pairs are given below.

\small
\begin{tabular}{p{0.45\textwidth}}
2010/01/08, I had lunch. I ate Indian food.\\
2010/01/09, I had cereals for breakfast with Hazel, Rylee, Piper, Nora, Avery, Eva, Nevaeh, Claire, Lydia, Olivia, Layla, Kinsley.\\
2010/01/09, I had lunch. I ate sushi.\\
2010/01/09, I had chinese food for dinner with Kayden, Carter.\\
2010/01/09, I spent 21 minutes on social media today.\\
2010/01/10, I did some hiking on 2010/01/10.\\
2010/01/10, I ate pasta for dinner.\\
2010/01/11, I talked to Nevaeh, Piper, Olivia, Eva for 37 minutes late in the evening.\\
2010/01/12, I did some swimming on 2010/01/12.\\
2010/01/12, I talked to Nora for 47 minutes in the morning.
:
\end{tabular}
\normalsize

\subsubsection{Example question-answer pairs}

\noindent
{\bf Atomic QA pairs:}  These QA pairs are created as each episode in the timeline is generated. Based on the episode that is generated, a question is instantiated from a set of templates and the answer to the question is extracted from the generated episode. Some examples are shown below.

\medskip
\noindent
\small
\begin{tabular}{p{0.45\textwidth}}
\em Q: What did I eat with Kayden and Carter on 2010/01/09? \\
\em A: I ate chinese food with Kayden and Carter.\\
\em Q: How long did I talk to Nora on 2010/01/12?\\
\em A: I talked to Nora for 47 minutes.
\end{tabular}
\normalsize

\medskip
\noindent
{\bf Complex QA pairs:} Using our query templates, we created 42
complex questions in our benchmark for the subset of categories we have implemented in our timeline generator. The answers are computed by applying external algorithms (e.g., SQL queries) over the timeline.

\medskip
\noindent
\small
\begin{tabular}{p{0.45\textwidth}}
\em Q: How much time on average did I spend on reading the news each day? \\
\em A: On average, you spent 32 minutes reading the news each day. \\
\em Q: How many times did I take my kids to an optician in 2010? \\
\em A: You took your kids 2 times to an optician. 
\end{tabular}
\normalsize

\section{Baselines and experimental results}

\subsection{Datasets}

Table \ref{tab:stat} summarizes the lifelogs we generate for \bname.
The dataset consists of 128M lifelog entries in total 
for all 3 types of densities (sparse, medium, and dense). Each entry
has an average of 8.4 tokens. \bname covers 25 categories of events
ranging from daily chat to lifetime events such as college graduation.
Different categories occur at various frequencies and 
describe events in heterogeneous formats at various lengths.
See Table \ref{tab:statfull} in the appendix for the full breakdown.
For our QA experiment, we uniformly sample 40 lifelogs for each density
(120 in total) as the hold-out test set.

\begin{table}[!t]
\centering
\caption{\small Statistics of 1,000 sparse, medium, and dense lifelogs. 
See Table \ref{tab:statfull} for the breakdown on the 25 event categories.}
\label{tab:stat}
\small
\begin{tabular}{cccc} \toprule
Datasets & \#Logs &  \#Entries   & Avg. \#Tokens \\ \midrule
sparse   & 1,000  & 14,941,703  & 8.51     \\
medium   & 1,000  & 34,522,030  & 8.12     \\
dense    & 1,000  & 78,559,743  & 8.50     \\ \midrule
all      & 3,000  & 128,023,476 & 8.40      \\ \bottomrule
\end{tabular}
\vspace{-2mm}
\end{table}

\setlength{\tabcolsep}{2pt}
\begin{table}[!t]
\small
\centering
\caption{\small Statistics of multi-hop QA tasks. } \label{tab:multihopdata}
\vspace{-2mm}
\begin{tabular}{cccccc}
\toprule
      & \#Logs & \#QA's & \#Evidence & AVG & \%Truncated \\ \midrule
Train & 240         & 8,586       & 10M        & 1,174.8            & 20.44\%     \\
Valid & 120         & 4,302       & 5M         & 1,216.6            & 20.99\%     \\
Test  & 120         & 4,284       & 5M         & 1,169.8            & 20.40\% \\ \bottomrule
\end{tabular}
\vspace{-2mm}
\end{table}

For each lifelog, we construct test samples for both \emph{atomic QA} and
\emph{multi-hop QA}. Atomic QA refers to the 
\textit{what, where, when, yes/no} types of questions where the answer requires reasoning 
(or plain extraction) over a valid span of a single input episode. 
We construct 5,000 such questions for each lifelog (600k in total) 
as the hold-out test set.
Multi-hop QA refers to the complex type of questions that involve selection and aggregation.

Table \ref{tab:multihopdata} shows the statistics of
the multi-hop QA datasets. In addition to the test set, 
we constructed a disjoint training and validation
set similarly (240 and 120 logs, respectively) for our fine-tuning experiment. 
Each lifelog contains $\sim$35 multi-hop queries. Each query also comes
with a set of ground-truth evidence records, which are all 
the episodes for deriving the correct answers. Each question
has an average of $>$1k evidence records, which together are
beyond the typical max length of 512/1024 tokens of transformer-based LMs. 
Indeed, even as we set the max input
length to 1024, $\sim$20\% of the input episodes are
truncated.

\subsection{Atomic QA}

We consider the following QA implementations for atomic QA:

\textbf{RAG} \citep{lewis2020rag}. This is a retrieval-augmented generative QA system, where we first retrieve some documents based on the query, and then condition the answer generator based on these retrieved documents and the query. We replace the Wikipedia based memory in the original RAG with episodes. We use the original \textit{RAG-Token} model released checkpoints.\footnote{See the implementation in \url{https://haystack.deepset.ai/tutorials/07_rag_generator}}

\textbf{ExtractiveQA} \citep{karpukhin-etal-2020-dense}. The key difference from RAG, is that the answering system is a span-based extractive model, extracting the answer from a given context. Specifically, the reader is a RoBERTa \citep{liu2019roberta} model fine-tuned on SQuAD~\citep{rajpurkar2018know}.\footnote{Details available at \url{https://huggingface.co/deepset/roberta-base-squad2}}


In both cases, we encode all the episodes using a dense passage retriever, and use FAISS to return the top-5 episodes. The retrieved documents are then fed into the answering component, and we get the top-1 answer. We consider 3 different setups for the retriever: Zero-shot (ZS) using the pre-trained checkpoints, fine-tuned on question-episode pairs from the lifelogs (FT), and oracle retrieval (OR) where we use the ground-truth episode associated with the question.
\begin{table}[!t]
\caption{\small Atomic QA Results comparing extractive and RAG based QA under 3 conditions for the retriever: Zero-shot (ZS), fine-tuned (FT), and oracle (OR).}
\vspace{-2mm}
\label{tab:atomicqa}
\small
\centering
\begin{tabular}{llrr}
\toprule
  Pipeline & Retriever &  Exact Match &       F1 \\
\midrule
Extractive &        FT &     82.6 & 93.8 \\
Extractive &        OR &     83.3 & 94.8 \\
Extractive &        ZS &     24.1 & 47.3 \\ \midrule
       RAG &        FT &     40.3 & 57.5 \\
       RAG &        OR &     73.7 & 84.4 \\
       RAG &        ZS &     8.4 & 32.9 \\
\bottomrule
\end{tabular}
\vspace{-2mm}
\end{table}

From the results in Table~\ref{tab:atomicqa}, we observe that extractive QA performs significantly better than generative QA, which is to be expected, given the benchmark construction, where the answers are always a valid span in the input for atomic queries. Furthermore, by fine-tuning the retrievers on the episodic data, we get a significant boost in performance for both extractive and rag setups, indicating that the QA systems do not generalize well to episodic data, and that improving retrieval is crucial to getting good performance from these models, particularly for RAG. After fine-tuning, the generative model performance still lags behind the extractive setup. 

\subsection{Multi-hop QA}

\setlength{\tabcolsep}{3pt}
\begin{table}[!t]
\small
\centering
\caption{\small Zeroshot (ZS) / Finetuned (FT) model performance on 
multi-hop QA over 120 \bname lifelogs.} \label{tab:multihop}
\vspace{-2mm}
\begin{tabular}{cc|cc|cc|cc}
\toprule
\multicolumn{2}{c|}{Retriever }                  & \multicolumn{2}{c|}{Oracle } & \multicolumn{2}{c|}{FT-retriever} & \multicolumn{2}{c}{ZS-retriever } \\
            
Reader      & size & ZS         & FT         & ZS           & FT          & ZS            & FT           \\ \midrule
Tapex-base  & 140M & 2.8 & 57.7 & 2.7 & 30.8 & 2.7 & 30.7 \\
Tapex-large & 400M & \textbf{6.5} & \textbf{59.0} & 6.5 & 32.7 & 6.5 & 33.0 \\
Bart-base   & 140M & 0.0 & 54.4 & 0.0 & 28.7 & 0.0 & 29.1 \\
Bart-large  & 400M & 0.0 & 47.0 & 0.0 & 21.9 & 0.0 & 25.2 \\ 
\bottomrule
\end{tabular}
\vspace{-2mm}
\end{table}

\begin{table}[!t]
\caption{\small Breakdown of Tapex-large (finetuned) with oracle 
retriever on question types and sizes of evidence sets.}
\vspace{-2mm}
\label{tab:multihopbreakdown}
\small
\centering
\begin{tabular}{ccc|ccc}
\toprule
type    & accuracy & total & \#evidence        & accuracy & total \\
\midrule
average & 11.1     & 360   & {[}0, 10{]}        & 85.1     & 1,949  \\
count   & 75.9     & 1,776  & (10, 100{]}        & 52.5     & 1,275  \\
argmax  & 47.2     & 1,668  & (100, 1000{]}      & 19.2     & 689   \\
list    & 62.7     & 480   & \textgreater{}1000 & 4.3      & 371  
\\ \bottomrule
\end{tabular}
\vspace{-5mm}
\end{table}

\begin{table*}[!t]
\small
\caption{\small Example correct and incorrect model predictions for multi-hop questions.} \label{tab:multihopexample} \vspace{-2mm}
\resizebox{.98\textwidth}{!}{
\begin{tabular}{p{4.5cm}ccp{7cm}} \toprule
Questions                                                       & groundtruth               & prediction                     & Notes                                                                                                       \\ \midrule
How long do I spend on average each day talking to my friends? & 84.05                     & \textcolor{myred}{83.94}                    & The question requires aggregating a total of 74k records                                                    \\ \midrule
In what year did I buy facial wash the most?                   & 2006                      & \textcolor{myred}{2015}                      & This question only needs to deal with 47 records, but requires complex arithmetic reasoning (count+compare) \\ \midrule
How many times did I have tacos for dinner in September 2019?  & 5                         & \textcolor{mygreen}{5}                         & The model correctly captures simple counting (5 evidence records)                                           \\ \midrule
Which places in New York, US did I visit with  Sofia?          & {[}'Central Park', ...{]} & \textcolor{mygreen}{{[}'Central Park', ...{]}} & The model correctly selects the 7 relevant locations from the input table                                  \\ \bottomrule
\end{tabular}}
\vspace{-3mm}
\end{table*}


Given the task's nature of aggregating structured data, 
we consider a baseline
based on TableQA~\cite{badaro2023transformers}.
In short, a table QA model answers questions by taking as input
a relational table (e.g., records of dental visits)
and a NL query.

We constructed the tables for table QA using an information extraction pipeline over the episodes as they are generated. By exploiting the topics (e.g., medical care, chat, exercise) which are known to the generation pipeline, we define a fixed schema for each topic. For example, we use the schema {\tt (date, place, medical\_care\_type, person)} for all types of medical care episodes, and run named-entity recognition to extract the tuple from each episode. For example, the record {\ tt (2019/03/23, annual vision checkup, university hospital, Jack) will be created from the input ``{\em I took Jack for his/her for an annual vision checkup on 2019/03/23 at the university hospital.} We then form the ``annual\_medical\_care'' table using all quadruples extracted from episodes under the same topic. 
This simple pipeline works very well (near perfect) for by exploiting the generation pipeline. For real-life lifelog data, additional challenges such as episode construction, topic/attribute discovery, and schema reconciliation, are beyond our current scope.

Due to the large size of the life logs that cannot fit
in the max length of LMs, 
the TableQA baseline also leverages a dense \emph{retriever}
for retrieving relevant records and constructing a concise
table representation of the entries. 
We then apply the TableQA model as the \emph{reader}
to produce the final answer via selection, aggregates, etc.

More precisely, for multi-hop queries, 
given a question $q$ over a set of life logs 
$L = \{l_1, \dots, l_n\}$, the retriever is a model $M_{\mathsf{ret}}$
where $L_{\mathsf{ret}} = M_{\mathsf{ret}}(q, L) \subseteq L$ is
the retrieved subset. We then process $L_{\mathsf{ret}}$ into
table format via NER and pattern matching to convert $L_{\mathsf{ret}}$ into
its table representation $T_{\mathsf{ret}}$.
Finally, the TableQA model $M_{\mathsf{read}}$
returns the answer $M_{\mathsf{read}}(q, T_{\mathsf{ret}})$.


We also evaluate variants of the 
Tapex~\cite{DBLP:conf/iclr/LiuCGZLCL22} model
as baselines. 
Tapex achieved the state-of-the-art performance 
of TableQA by pre-training a seq-to-seq LM on 
table datasets to mimic the behavior of a SQL engine.
We also compare the performance of Tapex with BART~\cite{DBLP:conf/acl/LewisLGGMLSZ20},
which has the same architecture as Tapex but without 
training on tabular data. For both models, we evaluate using the
denotation accuracy as in standard TableQA 
tasks~\cite{zhongSeq2SQL2017}. We evaluate each model
under both the zero-shot setting and with fine-tuning on the 
training sets. 
We also test InstructGPT as a baseline large LM, but leave the full result
in Table \ref{tab:multihop-gpt} in the appendix due to limited space.

Similar to atomic QA,
we evaluate each model under 3 settings of retrievers.
We first assume an \emph{oracle} retriever which has access
to the ground truth set of evidence to construct the input table.
A \emph{zero-shot} retriever uses a set of user-defined patterns
such as \emph{``I talked to X for Y minutes''} to find matching episodes 
(the same set of rules for converting episodes to table records).
We uniformly sample episodes up to the max length of the LM.
A \emph{fine-tuned} retriever trains a dense retriever model~\cite{DBLP:conf/emnlp/ReimersG19} 
from the training set and returns episodes closest to the question's dense embedding.

Table \ref{tab:multihop} summarizes the results. 
Overall, the 400M-parameter Tapex model achieves the best result
with fine-tuning and the oracle retriever. The 59\% accuracy is also
close to the Tapex's performance on the WikiTableQuestions benchmark~\cite{DBLP:conf/iclr/LiuCGZLCL22}.
However, its performance greatly reduces (1) under the zero-shot setting (6.5\%) or
(2) with a non-oracle retriever (33\%). Tapex generally outperforms its counterpart BART, which indicates the importance of understanding structured data and
aggregation for the multi-hop tasks. We also notice that fine-tuning the retriever generally does not improve the QA performance.
This can be due to the hard requirement of retrieving the exact evidence set
to correctly answer certain questions like count and average.

\subsection{Error analysis}

Table \ref{tab:multihopbreakdown} shows the breakdown of Tapex-large's
fine-tuning performance with a perfect retriever. Among the 4 types of questions,
argmax and average have the worst performance, likely because
they require arithmetic reasoning. We also observe that the model accuracy
decreases significantly (from 85.1\% to 4.3\%) as the number of evidence records grows,
which indicates the hardness of dealing with large input tables.
Table \ref{tab:multihopexample} shows examples of (in)correct
predictions of the model.



\section{Conclusions}
We presented \bname, a benchmark for generating lifelogs of imaginary people. Our experiments, with  state-of-the-art QA models, showed that there is still significant room for improving QA  over lifelog data. Specifically, while extractive systems can achieve impressive performance on \bname{} for atomic queries, the best performing QA system for multi-hop queries scores only 59.0\% in the perfect setting where the ground truth set of episodes are available.

We view the current state of \bname\ as a first version that will be enhanced in several ways as the QA technology improves. In future enhancements the episodes can be made more realistic and varied to also include events such as driving one's children to practices, or car breakdowns, to more unexpected events such as experiencing an earthquake etc. In addition, episodes can be enhanced to include different modalities, such as photos or videos of the episodes and more complicated queries can be included such as  ``{\em How many times did I swim in the month before I traveled to Machu Picchu?}''. 
Ideally, with appropriate obfuscations to preserve privacy, a future version can mirror precisely the lifelogs of real people. 


\section{Limitations and Ethical Considerations}

There are several perspectives from which we need to consider the ethical considerations of this work.

\medskip
\noindent
{\bf Privacy:} Lifelogs are personal data and should only be used and shared given user authorization. The lifelogs presented here are {\em fictitious} and do not reveal the personal information of any individual. No personal data is used to create this benchmark. This work is intended to unlock development in the creation, maintenance, querying and usage of lifelogs, and additional work will certainly be needed to ensure that they are secure and being meaningfully and responsibly used.  

\smallskip
\noindent
{\bf Comprehensiveness and diversity:}
We recognize that the lifelogs generated in this work are far from representing the full range of human experiences. While we strived to make the lifelogs complex enough to benchmark and compare current state-of-the-art, these lifelogs would not be considered diverse in the sense that a social scientist would note, and are likely biased by the life experiences of its creators. 
We encourage future work in creating lifelogs that are more inclusive and faithful to all walks of life. This includes further work in making lifelogs that are more diverse in terms of life experiences, personas, time scales, and queries as well as more granular and complex in detail. The strength of the benchmark is in identifying {\em patterns} of questions on lifelogs rather than the specific events described in them. 

\smallskip
\noindent
{\bf Inferring episodes:}
\bname\ is a collection of time-and-space boxed episodes, and not the raw data itself from which the episodes are inferred (e.g., a wedding photo, or video snippet from smart glasses). Naturally, more research would need to be devoted to understanding how to extract important information in natural language and infer episodic events from this raw data before performing question answering. As mentioned previously, this also involves sometimes grappling with the linguistic variation amongst the language used in the episode description and the query itself.

\smallskip
\noindent
{\bf Intended use:}
We clarify that the benchmark should not be used to train models for making key decisions that will impact people's lives (e.g., job matching, insurance approvals or building personal assistants). The intended use of \bname{} is as a benchmark to reveal potential limitations of QA systems over lifelog data. 
Even if the benchmark is determined to be sufficiently comprehensive, a detailed study should be conducted to understand the potential representational harms of using \bname{} before using it for training models. Conceivably, \bname{} can also facilitate research in evaluating the biases of QA systems by creating counterfactual pairs in the dataset: two timelines which are exactly the same, but differ by the demographic group or a specific life event (e.g., having dropped out of college or committed a crime). The QA system can then be systematically probed for differences in performance between the two timelines.


\bibliographystyle{acl_natbib}
 \bibliography{anthology,epiben}

\clearpage
\appendix

\section{Benchmark Statistics}

\subsection{Categories of questions}
\label{sec:categories}
The crowdsourced questions from 7 people led to the  categories of questions shown in Table~\ref{tbl:question_categories}. We gave 7 people the task of writing down questions over their potential lifelogs, and also categories of their questions. We than merge the categories which resulted in the categories shown in Table~\ref{tbl:question_categories} below. 

\begin{table*}[!h]
\caption{Categories of questions and some examples}
\label{tbl:question_categories}
\small
\begin{tabular}{c|p{7cm}|p{6cm}}
\toprule
 Episode Category    &  Explanation & Example queries \\
 \midrule
Care for oneself & Preventive medical appointments, self-care (e.g., massages, pedicures), medications, health metrics (e.g., heart rate, blood pressure) & When was the last time I visited my dentist? What was my average heart rate last week?\\
Taking care of parents & Visiting parents or family gatherings, taking them for health checkups and self-care, administering medications
& When was the last time I took my dad for his annual checkup?When was the last time I had dinner with my parents?\\
Raising children & Celebrating milestones, taking them for checkups/vaccinations, special moments & When was the last time my child had her yearly checkup? What type of cake did we buy for her last birthday?\\
Pets & First time pet arrived, pet's birthday, pet care/grooming, loss of pet & When was the last time my pet was groomed? How much did I spend on pet care last year? When did my pet pass away? \\
Accidents and recovery & Details of accidents, experiences, and recovery & How old was I when I fell from my bike? How many stitches did I receive from my bike accident? \\
Socializing & Spending time with friend, party, memorable conversations, dating, celebrations of events/holidays & How often did I chat with Avery last year? When was the first time I met Avery?\\
Daily life & Eating, cooking, drinking, shopping, religious practice, exercising, walking, meditating & When was the last time I visited restaurant X? How often did I cook pasta last month? How long did I meditate last week? \\
Entertainment & Hobbies, watching sports, participating in sports, watching media, reading media & How long did I exercise last week? when did I first learn to play the piano? where is the meditation group to meet this week? who went to watch the fashion show with me last Friday? \\
Life Milestones & Starting and graduating from schools, interviewing for jobs, starting and quitting jobs, promotions, engagement, marriage and divorce, anniversaries, work milestones, enrichment activities
& When was my first job interview? 
Where did we go for the anniversary last year? \\
Managing Finances & Investment decisions, credit score tracking	& 
How much did my daughter obtain from the trust last year? how much did I pay for my first investment property?\\
Travel & Travel preparation, getting there (by air, water, car), events during travel &	Did I take any photo in front of Big Ben?
		Are we going to London from the hotel by car?
		How much did the airbnb total for our last London trip?\\
Housing	& Finding a place to live, housework, house maintenance & When did I move the last time?
		did I make an appointment to clean the drains?
		when did I last purchase the laundry pods?\\
Diary Entries / Journaling	& Anything I may want to remember about my day, the conversations I had or other experiences I've gone through &
I went to a friend's graduation ceremony. Interesting conversation with a stranger at a grocery store. \\
\hline
\end{tabular}
\end{table*}

\subsection{Events}
\begin{table*}[!h]
\caption{Breakdown of \bname\ by events.}
\label{tab:statfull}
\small
\begin{tabular}{ccccccccc} \toprule
Event     & \#entries (M) & \#tokens & Category           & \#entries (M) & \#tokens & Category                       & \#entries                           & \#tokens                     \\ \midrule
chat         & 40.76         & 11.19    & hobbies            & 2.39          & 6.05     & birth\_info                    & 3,000                               & 8.23                         \\
watch tv     & 17.77         & 7.25     & dining             & 1.25          & 15.69    & college move                   & 726                                 & 10.62                        \\
read         & 11.87         & 5.00     & pet care           & 0.72          & 6.00     & college graduation             & 726                                 & 11.60                        \\
breakfast    & 9.56          & 6.89     & places visited     & 0.70          & 13.64    & grad school move               & 3                                   & 11.00                        \\
dinner       & 9.56          & 6.17     & bake               & 0.41          & 16.94    & grad school graduation         & 3                                   & 8.00                         \\
lunch        & 9.55          & 6.17     & cook               & 0.41          & 15.72    & \cellcolor[HTML]{C0C0C0} \textbf{Summary}                               & \cellcolor[HTML]{C0C0C0}                                    & \cellcolor[HTML]{C0C0C0}                             \\
exercise     & 8.99          & 3.17     & child med. care    & 0.22          & 15.91    & \cellcolor[HTML]{C0C0C0}sparse & \cellcolor[HTML]{C0C0C0}14,941,703  & \cellcolor[HTML]{C0C0C0}8.51 \\
social media & 5.93          & 6.00     & travel             & 0.17          & 10.79    & \cellcolor[HTML]{C0C0C0}medium & \cellcolor[HTML]{C0C0C0}34,522,030  & \cellcolor[HTML]{C0C0C0}8.12 \\
grocery      & 4.78          & 18.94    & personal med. care & 0.16          & 11.40    & \cellcolor[HTML]{C0C0C0}dense  & \cellcolor[HTML]{C0C0C0}78,559,743  & \cellcolor[HTML]{C0C0C0}8.50 \\
dating       & 2.64          & 8.00     & parent med. care   & 0.16          & 15.90    & \cellcolor[HTML]{C0C0C0}all    & \cellcolor[HTML]{C0C0C0}128,023,476 & \cellcolor[HTML]{C0C0C0}8.40 \\ \midrule
\end{tabular}
\end{table*}

Table \ref{tab:statfull} summarizes the 25 main lifelog events in \bname.
Chat is the most frequent events with 40M occurrences in all the 3k lifelogs.
The grocery event tends to be longest event type since each entry not only 
describes the items purchases but also people met at shopping.
There are also rare events such as college / grad school moves and graduations
occurring with low probabilities.


\section{Fine-Tuning Setup}
\subsection{Atomic QA}
For fine-tuning the QA systems on the timeline episodes, we use haystack\footnote{\url{https://github.com/deepset-ai/haystack}} implementation for RAG and Extractive QA. For the retriever, we use ground truth training episodes in the training split, and then fine-tune\footnote{For detailed steps, follow the tutorial at \url{https://haystack.deepset.ai/tutorials/09_dpr_training}}using in-batch examples as hard negatives, with a batch size of $64$, learning rate of $1.5e-5$, weight decay $0.75$, and number of warmup steps $200$, for $1$ epoch. For the reader, we start with a fine-tuned ROBERTA model\footnote{\url{https://huggingface.co/deepset/roberta-base-squad2}}, with a batch size of $128$, warmup proportion of $0.2$, learning rate of $1e-5$, for $2$ epochs.

\subsection{Multi-hop QA}

Our implementation of multi-hop QA is based on 
the Tapex implementation in HuggingFace's Transformers library.\footnote{See \url{https://github.com/huggingface/transformers/tree/main/examples/research_projects/tapex}.}
We experimented with both the BART-base and Bart-large architecture with
or without table pre-training. For fine-tuning, we use a learning rate of
3e-5 with weight decay 1e-2, a batch size of 8, and a beam size of 5 for beam search decoding.
We set the max length of the input sequence (the serialized table) to 1,024 sub-word tokens
and the max length of the decoded response to 128 sub-word tokens.

Our multi-hop QA dense retriever implementation is based on the SentenceTransformers library (\url{https://www.sbert.net/}). We used the \textsf{all-MiniLM-L6-v2} model checkpoint
for the zero-shot setting. For fine-tuning, we randomly sample 20 true positive examples
from the grounth truth evidence list for every question in the training set as the
positive question-evidence pairs. We create the set of negative pairs by
randomly sampling question-evidence pairs where the question and evidence are
from different episode category (e.g., chat vs. dining), so that they are guaranteed hard negatives.
We fine-tune the model with a batch size of 16 and a learning rate of 3e-5.

We ran all experiments on an AWS \textsf{p4d} server with A100 GPU's 
(1 GPU is used for each run). The experiments took a total of 25.4 GPU hours.

\section{Multi-hop QA with InstructGPT}

Since large pre-trained LMs (LLMs) have shown promising 
zero-shot performance across QA tasks, we also
test the 175B-parameter
InstructGPT~\cite{DBLP:journals/corr/abs-2203-02155} on 100 sampled 
multi-hop \bname\ questions. 
Similar to to the experiments for TableQA, we leverage
3 settings of the retrievers: oracle, fine-tuned (FT), or zero-shot (ZS).
Because the model may generate free-form answers, 
we compute the accuracy by manually checking whether 
the answers are compatible with the ground truth.
As such, the numbers
are not directly comparable to those for TableQA.

As shown in Table \ref{tab:multihop-gpt},
InstructGPT signicantly outperforms TableQA readers in the zeroshot settings
(e.g., 33\% vs. 6.5\% accuracy). However, the performance still does
not outperform that of fine-tuned TableQA models (59\% accuracy).
The result suggests a potential direction of leveraging fine-tuned LLMs
for the \bname tasks.

\setlength{\tabcolsep}{3pt}
\begin{table}[htb]
\small
\centering
\caption{\small InstructGPT performance Results on multi-hop QA
We report the results on a sample of 100 questions.} \label{tab:multihop-gpt}
\begin{tabular}{cccc}
\toprule
Retriever                   & {Oracle } & {FT-retriever} & {ZS-retriever } \\
             \midrule
InstructGPT & 33.0  & 25.0  & 18.0  \\ \bottomrule
\end{tabular}
\end{table}

\end{document}